\pdfoutput=1

\documentclass[11pt]{article}

\PassOptionsToPackage{table}{xcolor}
\usepackage[final]{acl}

\usepackage{times}
\usepackage{latexsym}

\usepackage[T1]{fontenc}

\usepackage[utf8]{inputenc}

\usepackage{microtype}

\usepackage{inconsolata}

\usepackage{graphicx}

\usepackage{threeparttable}
\usepackage{booktabs}
\usepackage{amsmath}
\usepackage{fancyvrb}
\usepackage{xspace}
\usepackage{tcolorbox}
\usepackage{caption}
\usepackage{hyperref}
\usepackage{multirow}
\usepackage{array}
\usepackage{amssymb}
\usepackage{threeparttable}

\definecolor{lightgray}{gray}{0.95}
\newcommand{\placeholder}[1]{\textcolor{blue}{\texttt{#1}}}

\newtcolorbox{promptbox}[1][]{
  colback=lightgray,
  colframe=black!10,
  fontupper=\ttfamily\small,
  coltitle=black,
  boxrule=0pt,
  arc=2pt,
  outer arc=2pt,
  boxsep=5pt,
  left=8pt,
  right=8pt,
  top=5pt,
  bottom=5pt,
  title=#1
}

\usepackage{etoolbox}

\newif\ifcomments

\commentstrue %

\newcommand{\clear}{\textsc{CLEAR}\xspace}

\ifcomments
    \newcommand\cc[1]{\textcolor{blue}{[Chacha: #1]}}
      
    \newcommand{\chenhao}[1]{\textcolor{magenta}{\textsc{#1 ---CT}}}
    \else
    \newcommand\cc[1]{}
    \newcommand{\chenhao}[1]{}
    
\fi

\newcommand{\para}[1]{\paragraph{#1}}

\title{CLEAR: A Clinically-Grounded Tabular Framework for \\Radiology Report Evaluation}

\author{
 \textbf{Yuyang Jiang\textsuperscript{1}\thanks{Department of Statistics, University of Chicago}}, 
 \textbf{Chacha Chen\textsuperscript{1}\thanks{Department of Computer Science, University of Chicago}}, 
 \textbf{Shengyuan Wang\textsuperscript{2}\thanks{Department of Computer Science and Technology, Tsinghua University}}, 
 \textbf{Feng Li\textsuperscript{1}\thanks{Department of Radiology, University of Chicago}}, 
 \textbf{Zecong Tang\textsuperscript{3}\thanks{College of Control Science and Engineering, Zhejiang University}}, \\
 \textbf{Benjamin M. Mervak\textsuperscript{4}\thanks{Department of Radiology, University of Michigan}}, 
 \textbf{Lydia Chelala\textsuperscript{1}\footnotemark[4]}, 
 \textbf{Christopher M. Straus\textsuperscript{1}\footnotemark[4]}, 
 \textbf{Reve Chahine\textsuperscript{4}\footnotemark[6]}, \\
 \textbf{Samuel G. Armato III\textsuperscript{1}\footnotemark[4]\thanks{Co-senior authorship}}, 
 \textbf{Chenhao Tan\textsuperscript{1}\footnotemark[2]\footnotemark[7]} \\[0.5em]
 \textsuperscript{1}University of Chicago \quad
 \textsuperscript{2}Tsinghua University \quad
 \textsuperscript{3}Zhejiang University \quad
 \textsuperscript{4}University of Michigan
}

\begin{document}

\maketitle

\begin{abstract}

Existing metrics often lack the granularity and interpretability to capture nuanced clinical differences between candidate and ground-truth radiology reports, resulting in suboptimal evaluation. 
We introduce a \textbf{Cl}inically-grounded tabular framework with \textbf{E}xpert-curated labels and \textbf{A}ttribute-level comparison for \textbf{R}adiology report evaluation (\textbf{\clear}).
CLEAR not only examines whether a report can accurately identify the presence or absence of medical conditions, but also assesses whether it can precisely describe each positively identified condition across five key attributes: \texttt{first occurrence}, \texttt{change}, \texttt{severity}, \texttt{descriptive location}, and \texttt{recommendation}. 
Compared to prior works, CLEAR’s multi-dimensional, 
attribute-level outputs enable a more comprehensive and clinically interpretable evaluation of report quality.
Additionally, to measure the clinical alignment of \clear, we collaborate with five board-certified radiologists to develop \textbf{CLEAR-Bench}, a dataset of 100 chest X-ray reports from MIMIC-CXR, annotated across 6 curated attributes and 13 CheXpert conditions.
Our experiments show that CLEAR achieves high accuracy in extracting clinical attributes and provides automated metrics that are strongly aligned with clinical judgment.

\end{abstract}

\section{Introduction}

Evaluation is becoming increasingly challenging in the era of large language models (LLMs). While models continue to hill-climb on benchmarks rapidly 
\citep{aiindex2025, openai2025gpt-4.5, athropic2025claude, tu2025conversational, mcduff2025towards},
it remains unclear whether these reported metrics match task-specific needs~\citep{ganguli2023challenges, rauh2024gaps, bedi2025test}.
In the context of radiology, the pursuit of generalist foundation models achieves promising progress 
\citep{Bannur2024MAIRA2GR, ZambranoChaves2025}, but do these ``appealing'' automated metrics truly capture clinically aligned qualities~\cite{paschali2025foundation}?

\begin{figure*}[t]
    \centering
    \includegraphics[width=\textwidth]{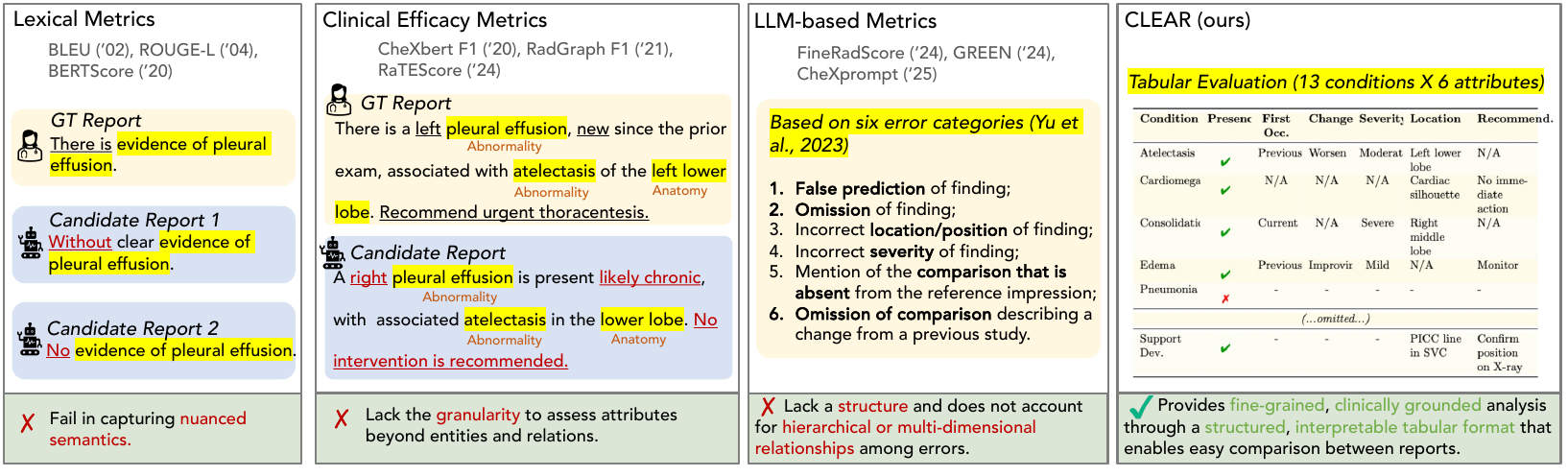}
    \caption{A comparison of existing metrics with CLEAR. Yellow highlights indicate the main evaluation mechanism for each type of metric. Red underlining marks an erroneous term in the candidate report, in contrast to the black underlined term in the ground-truth report, which the designed metric fails to evaluate.}
    \label{fig:metric}
\end{figure*}

In the existing literature, three main types of metrics have been proposed to assess the quality of generated radiology reports, as illustrated in \autoref{fig:metric}:
(i) \textbf{Lexical metrics} measure surface-level similarity between the generated and ground-truth reports \cite{papineni2002bleu,lin2004rouge,zhang2019bertscore}. While straightforward and easy to compute, they struggle to capture nuanced semantics and domain-specific terminology, leading to poor sensitivity to clinically significant errors.
(ii) \textbf{Clinical efficacy metrics} evaluate the correctness of medical entities and their relationships \cite{jain2021radgraph,yu2023evaluating,zhao-etal-2024-ratescore}, typically through structured extraction-based comparisons. Although more clinically informed than lexical metrics, they lack the resolution to assess fine-grained attributes such as severity, temporal progression, or treatment recommendations.
(iii) \textbf{LLM-based metrics}
\cite{ostmeier2024green,huang2024fineradscore,ZambranoChaves2025} represent the latest direction, often leveraging the pipeline of LLM-as-a-Judge~\cite{zheng2023judging} with pre-defined taxonomies such as the six error categories from ReXVal dataset~\cite{yu2023rexval}. While getting closer to expert judgment compared with previous two types, these methods may still lack comprehensive structured attribution and condition-level interpretability.

\begin{figure*}[t]
    \centering
    \includegraphics[width=\textwidth]{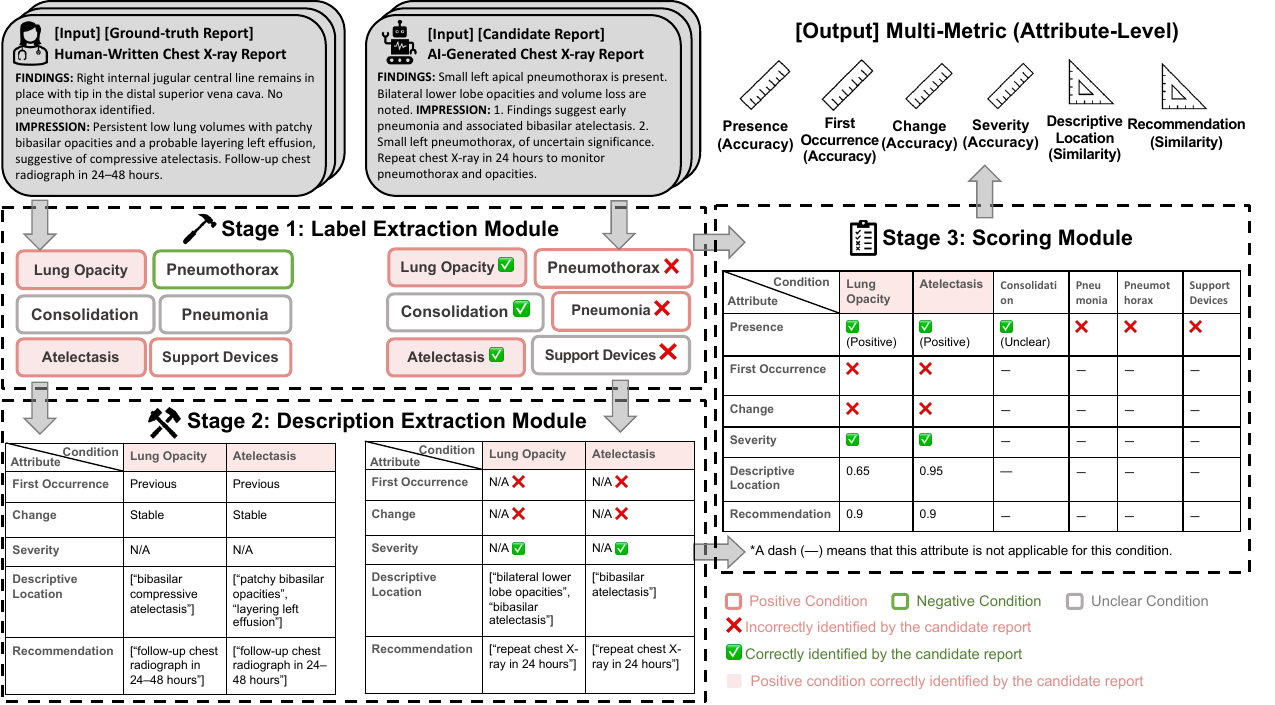}
    \caption{\textbf{CLEAR Framework.} Given a pair of ground-truth and candidate reports, we first assesses whether the candidate report can accurately identify a set of medical observations in the \textbf{label extraction module}. For each correctly identified positive condition, the \textbf{description extraction module} further evaluates the report’s ability to describe the condition across five attributes: \texttt{first occurrence}, \texttt{change}, \texttt{severity}, \texttt{descriptive location}, and \texttt{recommendation}. Finally, the \textbf{scoring module} compiles and outputs the evaluation metrics.}
    \label{fig:workflow}
\end{figure*}

Therefore, to address the limitations of existing metrics, we introduce \textbf{CLEAR} (Section~\ref{sec:pipeline}), the first  clinically-grounded attribute-level evaluation framework that leverages LLMs to map free-text radiology reports to a structured tabular format. Compared to prior work, CLEAR transforms the coarse, single-dimensional taxonomy into a fine-grained, multidimensional structure. Our design not only enables more comprehensive comparisons between candidate and ground-truth reports, but also provides interpretable outputs to assess report quality at the level of condition-attribute pairs.
Given the strong adaptability of LLMs across diverse language tasks, they serve as an ideal unified model to operationalize our proposed framework.

Specifically, CLEAR begins with the \textbf{Label Extraction Module} (Section~\ref{sec:module1}), which evaluates whether the candidate report can precisely identify the presence or absence of specific medical conditions. To ensure robust performance across model scales, we enhance this module using high-quality, expert-curated labels.
Next, for each correctly identified positive condition, the \textbf{Description Extraction Module} (Section~\ref{sec:module2}) assesses whether the candidate report can accurately describe the condition. Jointly established with one research radiologist and reviewed by one clinical radiologist, we define five commonly used attributes in a radiology report (\texttt{first occurrence}, \texttt{change}, \texttt{severity}, \texttt{descriptive location}, and \texttt{recommendation}), enabling the first systematic evaluation of these critical facets.
Finally, the \textbf{Scoring Module} (Section~\ref{sec:module3}) compiles and outputs metric scores for each attribute. We carefully design automated measurements based on the output type from previous modules: accuracy metrics aim at exact matches for single-label outputs while similarity metrics focus on contextual relevance for multi-phrasing outputs.

Additionally, since no existing datasets~\citep{tian2023refisco, yu2023rexval, rao2025rexerr} are compatible with CLEAR, we work closely with radiologists to create \textbf{CLEAR-Bench} (Section \ref{sec:clear-bench}), an expert-curated, attribute-level dataset to assess clinical alignment. 
CLEAR-Bench consists of 100 studies randomly sampled from MIMIC-CXR-JPG test and validation sets~\citep{johnson2019mimic,johnson2024mimic}. Each study is annotated and reviewed by at least two radiologists across 6 report attributes and 13 CheXpert conditions\footnote{Atelectasis, Cardiomegaly, Consolidation, Edema, Enlarged Cardiomediastinum, Fracture, Lung Lesion, Lung Opacity, Pleural Effusion, Pleural Other, Pneumonia, Pneumothorax, and Support Devices.}~\citep{irvin2019chexpert}.
CLEAR-Bench includes two components: 
(i) \textbf{Expert ensemble labels} includes ground-truth labels for \texttt{presence} attribute of each condition. These labels are constructed via majority voting among three radiologists, followed by one round of consensus discussion. 
(ii) \textbf{Expert curated attributes} contains the remaining five report attributes for each condition positively identified in the ensemble labels. 
These attributes are first generated by LLMs, then independently curated by two radiologists, and finalized through one round of discussion and resolution.
Additionally, during the curation process, we collect expert Likert scores for each model output, contributing to the assessment of how well proposed automated metrics align with clinical judgment.

Finally, we evaluate each component of CLEAR using the CLEAR-Bench. Our experimental results (Section~\ref{sec:experiment}) show that:
(i) the Label Extraction Module achieves high accuracy compared to expert ensemble labels and significantly outperforms existing labelers across all metrics;
(ii) the Description Extraction Module can accurately extract attribute-level information according to clinical assessment;
(iii) our proposed automated metrics serve as effective proxies for expert scoring.

\begin{table*}[t]
\small
\centering
\begin{threeparttable}
\begin{tabular}{@{}>{\arraybackslash}p{3.2cm}p{7.6cm}>{\arraybackslash}p{2.4cm}>{\arraybackslash}p{1.7cm}@{}}
\toprule
\textbf{Attribute} & \textbf{Value Set} & \textbf{NLP Task} & \textbf{Metric} \\
\midrule
Presence & \footnotesize $S_{1} \in$ \{"Positive", "Unclear", "Negative"\} & 
\begin{tabular}[t]{@{}c@{}}
Cls (\hyperref[fig:presence-prompt]{Prompt 1}) 
\end{tabular} & Accuracy \\
\midrule
\multicolumn{4}{l}{\textit{\textbf{Temporal Assessment}}} \\
First Occurrence & \footnotesize $S_{2}\in$ \{"Previous", "Current", "N/A"\} & 
\begin{tabular}[t]{@{}c@{}}
QA (\hyperref[fig:occurrence-prompt]{Prompt 2}) 
\end{tabular} & Accuracy\\
Change & 
\begin{tabular}[t]{@{}c@{}}
\footnotesize $S_{3}\in$ \{"Improving", "Stable", "Worsening", "Mixed", "N/A"\}
\end{tabular} & 
QA (\hyperref[fig:change-prompt]{Prompt 3}) & 
Accuracy\\
\midrule
\multicolumn{4}{l}{\textit{\textbf{Description Assessment}}} \\
Severity & \footnotesize $S_{4}\in$ \{"Severe", "Moderate", "Mild", "Mixed", "N/A"\} & 
QA (\hyperref[fig:severity-prompt]{Prompt 4}) & 
Accuracy \\
Descriptive Location & 
\begin{tabular}[t]{@{}l@{}}
\footnotesize $S_{5}=$ \{Entry\textsubscript{1}, ..., Entry\textsubscript{n}\} \\
\footnotesize (e.g., Entry\textsubscript{m} = "left mid lung atelectasis")
\end{tabular} & 
IE (\hyperref[fig:location-prompt]{Prompt 5}) & 
Similarity \\
\midrule
\multicolumn{4}{l}{\textit{\textbf{Treatment Assessment}}} \\
Recommendation & 
\begin{tabular}[t]{@{}l@{}}
\footnotesize $S_{6}=$ \{Entry\textsubscript{1}, ..., Entry\textsubscript{n}\}\\
\footnotesize (e.g., Entry\textsubscript{m} = "recommend follow-up at 4 weeks")
\end{tabular} & 
IE (\hyperref[fig:action-prompt]{Prompt 6}) & 
Similarity \\
\bottomrule
\end{tabular}
\begin{tablenotes}
\item$^*$ \footnotesize Cls denotes “Classification,” QA denotes “Question Answering,” and IE denotes “Information Extraction.”
\end{tablenotes}
\end{threeparttable}
\caption{An overview of our expert-curated fine-grained attributes in CLEAR.}
\label{tab:module}
\end{table*}

\section{CLEAR Framework}
\label{sec:pipeline}

We introduce the CLEAR framework, a hierarchical and fine-grained system for evaluating the clinical accuracy of radiology reports. CLEAR addresses both high-level diagnostic correctness and the descriptive quality of positive findings. As shown in \autoref{fig:workflow}, CLEAR includes three sequential stages: label extraction, description extraction, and structured scoring.

Specifically, given a ground-truth and a candidate report pair, CLEAR first identifies whether the candidate correctly recognizes the presence or absence of specific medical conditions (Stage 1). It then examines, for each positively identified condition, whether the ground-truth and candidate reports are aligned across a set of expert-curated descriptive dimensions (Stage 2). Finally, it aggregates these evaluations into standardized, multi-dimensional metrics (Stage 3).

\subsection{Stage 1: Label Extraction}
\label{sec:module1}

This stage determines the presence or absence of 13 pre-defined medical conditions in the candidate report, following the CheXpert structure~\cite{irvin2019chexpert}. 
Since accurately identifying and describing abnormalities is more clinically significant in radiology reporting, we exclude the ``No Findings'' label and focus on the remaining 13 conditions.
Each condition is labeled as \texttt{positive}, \texttt{unclear}, or \texttt{negative} based on report content.

While existing labelers like CheXbert~\cite{smit2020combining} and CheXpert~\cite{irvin2019chexpert} are available, our pilot analysis (see Table~\ref{tab:classification}) showed that their performance was limited. Since label extraction involves understanding and interpreting clinical narratives to assign structured labels, we hypothesized that LLMs could offer significant improvements over existing approaches. In particular, LLMs can handle complex linguistic nuances, such as negation, uncertainty, and context-dependent phrasing, more effectively in free-form radiology reports.

\para{Base model variants and training strategies.} We support three model scales: small (fine-tuned Qwen2.5-7B-Instruct and Llama-3.1-8B-Instruct), medium (Llama-3.3-70B-Instruct and Llama-3.1-70B-Instruct), and large (GPT-4o). For medium and large models, we apply different prompting strategies, including zero-shot (\hyperref[fig:presence-prompt]{Prompt 1}) and five-shot. For small models, we perform full-parameter fine-tuning using our curated dataset. To avoid overfitting, we first conduct hyperparameter tuning through 5-fold cross-validation and a grid search over learning rate, gradient accumulation steps, and number of epochs, followed by re-training on the full dataset.
Full implementation details are provided in Appendix~\ref{sec:model}.

\para{Expert-in-the-loop label curation.} 
High-quality labeled data is essential for training our label extraction model. To build a gold training dataset, we implemented a multi-stage annotation refinement with expert in the loop. 
We began with the test set from MIMIC-CXR-JPG~\citep{johnson2024mimic}, which includes a single radiologist's annotations for 13 CheXpert conditions~\citep{irvin2019chexpert}. Each condition is originally labeled as \texttt{positive}, \texttt{negative}, \texttt{unmentioned}, or \texttt{uncertain}.
In initial discussions with a radiologist, we identified two major issues with the original annotations: labeling errors (e.g., conditions mentioned in the report but left unlabeled) and category ambiguity (e.g., vague distinctions between \texttt{negative} and \texttt{unmentioned}).
To address these, we used GPT-4o to pre-screen and re-label the reports, prompting it with the original MIMIC labeling guidelines. We then flagged cases with label mismatches between GPT-4o and the original annotations.
We then asked an expert to re-annotate the discrepancy cases. 
To reduce the radiologist's workload, reports with more than five mismatched condition labels are discarded from expert annotation, as such extensive disagreement often signals deeper interpretive ambiguities or quality issues in the original reports. 
While this introduces potential bias, we prioritized curating a high-quality subset over exhaustively correcting all samples.
For the remaining reports, our collaborating radiologist independently re-annotated only the discrepant conditions, reviewing the original report text without seeing prior labels. 
During human annotation process, we observed that the original labeling schema lacked sufficient granularity to reflect the nuanced certainty levels expressed in radiology. In discussion with our expert radiologist, we expanded the label set to:\texttt{\{confidently present, likely present, neutral, likely absent, confidently absent\}}.
In total, we curated 550 studies, each with high-quality labels for all 13 conditions. For consistency with prior work and to simplify downstream modeling, we further merged all labels into three classes \{\texttt{positive, negative, unclear}\}. 
A detailed description of the annotation process and instructions are provided in Appendix~\ref{sec:data}.

\subsection{Stage 2: Description Extraction}
\label{sec:module2}

Building on the condition labels from Stage 1, this module extracts fine-grained clinical features that capture essential descriptive information for accurate reporting. 
The primary motivation is to transform the narrative text of radiology reports into a comprehensive, structured tabular format that distills all clinically significant attributes.
In collaboration with two radiologists, we developed five clinically significant dimensions: 
\texttt{first occurrence} (whether the condition is newly observed), 
\texttt{change} (progression or improvement from prior studies), 
\texttt{severity} (the extent or intensity of the condition), 
\texttt{descriptive location} (specific anatomical site), and 
\texttt{recommendation} (suggested follow-up actions).
These expert-developed attributes were specifically designed to reflect the nuanced but essential information radiologists routinely document when interpreting chest X-rays. 
By extracting these attributes, our approach enables a more comprehensive evaluation beyond simple condition detection.

\para{Implementation details.} 
We use prompt-based methods to extract each of the five attributes from free-text reports.
Each attribute can be naturally framed as a standalone language understanding task. To operationalize this, we design custom prompts tailored to the nature of each attribute: we use a Question Answering (QA) template to prompt the model for \texttt{first occurrence} (\hyperref[fig:occurrence-prompt]{Prompt 2}), \texttt{change} (\hyperref[fig:change-prompt]{Prompt 3}), and \texttt{severity} (\hyperref[fig:severity-prompt]{Prompt 4}), and an Information Extraction (IE) template for \texttt{descriptive location} (\hyperref[fig:location-prompt]{Prompt 5}) and \texttt{recommendation} (\hyperref[fig:action-prompt]{Prompt 6}). 
For QA tasks, the model  selects the best answer from multiple-choice options based on its understanding of the report. 
For IE tasks, it extracts relevant phrases guided by condition-specific example terminologies. 
Our prompt templates and terminology lists are summarized in Appendix~\ref{sec:prompt}, and were reviewed by two radiologists.
We use a single model to process all five prompt types, one prompt per query to extract each attribute from a given report. We evaluate two model scales: a smaller Llama-3.1-8B-Instruct and a larger GPT-4o from OpenAI.

\subsection{Stage 3: Scoring and Metrics}
\label{sec:module3}
In this module, we process outputs from Stage 1 and Stage 2 into numeric metrics for each attribute. 
Given the $i$-th pair of ground-truth and candidate attribute sets, denote the attributes extracted from the ground-truth report as $\{S_{j}^{(i)}\}_{j=1}^{6}$ and from the candidate report as $\{\hat{S}_{j}^{(i)}\}_{j=1}^{6}$. An overview of the attributes is provided in ~\autoref{tab:module}.

For \texttt{presence} $(S_1, \hat{S}_1)$, we evaluate the accuracy of identifying \texttt{Positive} and \texttt{Negative} conditions. We define a target class $c \in \{\texttt{Positive}, \texttt{Negative}\}$, treating all other labels as non-target. The corresponding binary F1 score, $F1_c$, is computed for each target class, resulting a positive-F1 and negative-F1. We report these scores at three levels: micro average, Top-5 condition average\footnote{Top five conditions in MIMIC-CXR-JPG are Pneumothorax, Pneumonia, Edema, Pleural Effusion, and Consolidation.}, and across all 13 conditions.

For \texttt{first occurrence} $(S_2, \hat{S}_2)$, \texttt{change} $(S_3, \hat{S}_3)$, and \texttt{severity} $(S_4, \hat{S}_4)$, we assess the exact match between predictions and ground truth. Considering that these attributes are framed as multiple-choice questions in the prompt, exact match is a natural and appropriate metric. Accuracy is calculated as $\text{Acc.}_{j} = \frac{\sum_{i} \mathbb{1}[S_{j}^{(i)} = \hat{S}_{j}^{(i)}]}{\sum_{i} 1}$. We report accuracy at the micro level, as well as averaged across reports and the 13 conditions.

For \texttt{descriptive location} $(S_5, \hat{S}_5)$ and \texttt{recommendation} $(S_6, \hat{S}_6)$, which involve free-text descriptions, we measure phrase-level similarity against clinically meaningful expressions. To evaluate alignment, we first use optimal matching–based metrics with similarity scores such as BLEU-4~\cite{papineni2002bleu} and ROUGE-L~\cite{lin2004rouge}:
\[
\text{Score}_{j}^{(i)} = \frac{1}{|S_{j}^{(i)}|} \sum_{e \in S_{j}^{(i)}} \max_{\hat{e} \in \hat{S}_{j}^{(i)}} \text{Similarity}(e, \hat{e}),
\]
where $S_{j}^{(i)} = \{e_k\}_{k=1}^{n}$ and $\hat{S}_{j}^{(i)} = \{\hat{e}_k\}_{k=1}^{n'}$.  
Additionally, to better approximate clinical judgment from an expert’s perspective, we prompt o1-mini (\hyperref[fig:o1-mini-prompt]{Prompt 8}) to directly compare each attribute pair and return a similarity score in the range $[0, 1]$.

\begin{table*}[t]
\small
\centering
\resizebox{\textwidth}{!}{
\begin{threeparttable}
\begin{tabular}{@{}p{6cm}@{\hspace{12pt}}c@{\hspace{12pt}}c@{\hspace{12pt}}c@{\hspace{12pt}}c@{\hspace{12pt}}c@{\hspace{12pt}}c@{}}
\toprule
\textbf{Experiments} & \textbf{Pos F1@13} & \textbf{Pos F1@5} & \textbf{Pos F1 (micro)} & \textbf{Neg F1@13} & \textbf{Neg F1@5} & \textbf{Neg F1 (micro)} \\
\midrule
\multicolumn{7}{c}{\textsc{Large Models}} \\
GPT-4o (base) & \textbf{0.805} & 0.929 & 0.934 & 0.476 & 0.648 & 0.815 \\
GPT-4o (5-shot) & 0.795 & \textbf{0.940} & \textbf{0.934} & 0.510 & 0.723 & 0.842 \\
\midrule
\multicolumn{7}{c}{\textsc{Medium Models}} \\
Llama-3.1-70B-Instruct (base) & 0.782 & 0.890 & 0.924 & 0.630 & 0.850 & 0.920 \\
Llama-3.1-70B-Instruct (5-shot) & 0.794 & 0.916 & 0.924 & \textbf{0.744} & 0.890 & \textbf{0.958} \\
Llama-3.3-70B-Instruct (base) & 0.780 & 0.894 & 0.925 & 0.602 & 0.876 & 0.926 \\
Llama-3.3-70B-Instruct (5-shot) & 0.781 & 0.907 & 0.926 & 0.695 & \textbf{0.892} & 0.953 \\
\midrule
\multicolumn{7}{c}{\textsc{Small Models}} \\
Llama-3.1-8B-Instruct (base) & 0.736 & 0.880 & 0.910 & 0.418 & 0.660 & 0.714 \\
Llama-3.1-8B-Instruct (550 finetune) & 0.729 & 0.806 & 0.905 & 0.482 & 0.803 & 0.949 \\
Qwen2.5-7B-Instruct (base) & 0.694 & 0.834 & 0.880 & 0.413 & 0.616 & 0.736 \\
Qwen2.5-7B-Instruct (550 finetune) & 0.727 & 0.800 & 0.905 & 0.511 & 0.849 & 0.953 \\
\midrule
\multicolumn{7}{c}{\textsc{Baselines}} \\
CheXbert~\cite{smit2020combining} & 0.695 & 0.833 & 0.897 & 0.498 & 0.877 & 0.952 \\
CheXpert~\cite{irvin2019chexpert} & 0.674 & 0.811 & 0.888 & 0.522 & 0.831 & 0.948 \\
\midrule
\textbf{$\Delta$ Improvement over SOTA}& +15.8\% & +12.8\% & +4.1\% & +42.5\% & +1.7\% & +0.06\%\\
\bottomrule
\end{tabular}
\end{threeparttable}
}
\caption{Evaluation of the label extraction module. CLEAR outperforms existing labelers across all metrics in identifying both positive and negative conditions. Specifically, larger models perform better at capturing positive conditions, while techniques such as 5-shot prompting and supervised fine-tuning significantly improve the detection of negative conditions. }
\label{tab:classification}
\end{table*}

\section{CLEAR-Bench: Attribute-Level Expert Alignment Dataset}
\label{sec:clear-bench}

In this section, we introduce CLEAR-Bench, an expert-curated, attribute-level dataset in collaboration with five radiologists. Inspired by recent expert evaluation
datasets for chest X-ray reports~\citep{tian2023refisco, yu2023rexval, rao2025rexerr}, CLEAR-Bench is specifically designed to assess how well automated evaluators like CLEAR align with radiologist judgments. It consists of two annotation subsets: expert ensemble labels and expert-curated attributes. We defer full details of the instruction criteria, interface design, and annotation workflow to Appendix~\ref{sec:data}.

\para{Expert ensemble labels.} 
These provide the ground-truth labels for the \texttt{Presence} attribute. We randomly selected 100 studies from the validation and test sets of MIMIC-CXR-JPG~\cite{johnson2024mimic}, excluding any training samples and normal studies. Each report was independently annotated from scratch by three board-certified radiologists. During annotation, the radiologists categorized each of 13 CheXpert conditions~\cite{irvin2019chexpert} into one of five categories: \texttt{confidently absent}, \texttt{likely absent}, \texttt{neutral}, \texttt{likely present}, and \texttt{confidently present}, based on their best interpretation of the report.
After the initial round of annotations, we merged \texttt{confidently present} and \texttt{likely present} into a single category \texttt{positive}, while \texttt{likely absent} and \texttt{confidently absent} into \texttt{negative}. We then assessed agreement across annotators. Remaining disagreements were first resolved by majority vote, followed by a consensus discussion for any unresolved conflicts. The finalized dataset serves as the ground truth for evaluating model performance in the Label Extraction Module.

\para{Expert-curated attributes.} These cover the remaining five report attributes: \texttt{first occurrence}, \texttt{change}, \texttt{severity}, \texttt{descriptive location}, and \texttt{recommendation}. 
We began by preparing two sets of model-generated attributes, one from Llama-3.1-8B-Instruct and the other from GPT-4o, for each positive condition identified in the expert ensemble labels. These two sets were merged and then randomly split into two review sets, each with 50 samples from Llama and 50 from GPT-4o. Each set was independently reviewed by separate radiologists.
During curation, each radiologist first rated each attribute as \texttt{incorrect}, \texttt{partially correct}, or \texttt{correct}. For non-\texttt{correct} attributes, the radiologist also provided a revised version, which was used to construct the ground-truth attribute set.

\section{Experiments}
\label{sec:experiment}

\para{Experimental setup.}
To evaluate the effectiveness and clinical reliability of our proposed CLEAR framework, we conduct experiments using CLEAR-Bench.
For the Label Extraction Module, we compare CLEAR’s performance against two established baselines: the BERT-based labeler CheXbert~\citep{smit2020combining} and the rule-based labeler CheXpert~\citep{irvin2019chexpert}, using the Expert Ensemble Labels from CLEAR-Bench. We report F1 scores as introduced in Section~\ref{sec:module3}.
For the Description Extraction Module, we evaluate CLEAR using the Expert-Curated Attributes from CLEAR-Bench. As no prior baselines exist for this task, we report expert evaluation scores directly, along with automated metrics defined in Section~\ref{sec:module3}.

\begin{table*}[t]

\centering
\footnotesize
\setlength{\tabcolsep}{3pt}  %
\renewcommand{\arraystretch}{1.1}  %
\begin{threeparttable}
\resizebox{\textwidth}{!}{
\begin{tabular}{@{}p{3.6cm}cc|cc|cc|cc|cc@{}}
\toprule
& \multicolumn{2}{c|}{\textbf{First Occurrence}} & \multicolumn{2}{c|}{\textbf{Change}} & \multicolumn{2}{c|}{\textbf{Severity}} & \multicolumn{2}{c|}{\textbf{Descriptive Location}} & \multicolumn{2}{c}{\textbf{Recommendation}} \\
\cmidrule(lr){2-3} \cmidrule(lr){4-5} \cmidrule(lr){6-7} \cmidrule(lr){8-9} \cmidrule(lr){10-11}
\textbf{Metric} & \textbf{GPT-4o} & \textbf{Llama 8B} & \textbf{GPT-4o} & \textbf{Llama 8B} & \textbf{GPT-4o} & \textbf{Llama 8B} & \textbf{GPT-4o} & \textbf{Llama 8B} & \textbf{GPT-4o} & \textbf{Llama 8B} \\
\midrule
\multicolumn{11}{c}{\textsc{Expert Evaluation Scores}} \\
 Experts (condition averaged) & \cellcolor{green!15}\textbf{0.818} & \cellcolor{yellow!15}0.685 & \cellcolor{green!15}0.837 & \cellcolor{yellow!15}0.685 & \cellcolor{green!15}\textbf{0.809} & \cellcolor{yellow!15}0.565 & \cellcolor{green!15}0.857 & \cellcolor{yellow!15}0.761 & \cellcolor{green!15}0.933 & \cellcolor{yellow!15}\textbf{0.474} \\

Experts (report averaged) & \cellcolor{green!15}0.783 & \cellcolor{yellow!15}\textbf{0.680} & \cellcolor{green!15}\textbf{0.867} & \cellcolor{yellow!15}\textbf{0.688} & \cellcolor{green!15}0.771 & \cellcolor{yellow!15}\textbf{0.583} & \cellcolor{green!15}0.872 & \cellcolor{yellow!15}\textbf{0.763} & \cellcolor{green!15}\textbf{0.940} & \cellcolor{yellow!15}0.416 \\

 Experts (micro) & \cellcolor{green!15}0.777 & \cellcolor{yellow!15}0.662 & \cellcolor{green!15}0.855 & \cellcolor{yellow!15}0.663 & \cellcolor{green!15}0.777 & \cellcolor{yellow!15}0.570 & \cellcolor{green!15}\textbf{0.867} & \cellcolor{yellow!15}0.757 & \cellcolor{green!15}0.936 & \cellcolor{yellow!15}0.404 \\
\midrule

\multicolumn{11}{c}{\textsc{Accuracy Metrics}} \\
Acc. (condition averaged) & \cellcolor{green!15}0.740 & \cellcolor{yellow!15}0.688 & \cellcolor{green!15}0.710 & \cellcolor{yellow!15}0.589 & \cellcolor{green!15}0.682 & \cellcolor{yellow!15}0.470 & -- & -- & -- & -- \\

 Acc. (report averaged) & \cellcolor{green!15}0.755 & \cellcolor{yellow!15}0.679 & \cellcolor{green!15}0.759 & \cellcolor{yellow!15}0.596 & \cellcolor{green!15}0.685 & \cellcolor{yellow!15}0.532 & -- & -- & -- & -- \\

Acc. (micro) & \cellcolor{green!15}0.737 & \cellcolor{yellow!15}0.665 & \cellcolor{green!15}0.754 & \cellcolor{yellow!15}0.575 & \cellcolor{green!15}0.671 & \cellcolor{yellow!15}0.494 & -- & -- & -- & -- \\
\midrule
\multicolumn{11}{c}{\textsc{Similarity Metrics}} \\
 o1-mini (micro) & -- & -- & -- & -- & -- & -- & \cellcolor{green!15}0.785 & \cellcolor{yellow!15}0.739 & \cellcolor{green!15}0.888 & \cellcolor{yellow!15}0.361 \\

ROUGE-L (micro) & -- & -- & -- & -- & -- & -- & \cellcolor{green!15}0.686 & \cellcolor{yellow!15}0.672 & \cellcolor{green!15}0.887 & \cellcolor{yellow!15}0.268 \\

 BLEU-4 (micro) & -- & -- & -- & -- & -- & -- & \cellcolor{green!15}0.500 & \cellcolor{yellow!15}0.402 & \cellcolor{green!15}0.885 & \cellcolor{yellow!15}0.263 \\
\midrule
\textbf{Average (experts)} & \cellcolor{green!25}0.793 & \cellcolor{yellow!25}0.676 & \cellcolor{green!25}0.853 & \cellcolor{yellow!25}0.679 & \cellcolor{green!25}0.786 & \cellcolor{yellow!25}0.573 & \cellcolor{green!25}0.865 & \cellcolor{yellow!25}0.760 & \cellcolor{green!25}0.936 & \cellcolor{yellow!25}0.431 \\

\textbf{Average (all)} & \cellcolor{green!25}0.768 & \cellcolor{yellow!25}0.677 & \cellcolor{green!25}0.797 & \cellcolor{yellow!25}0.633 & \cellcolor{green!25}0.733 & \cellcolor{yellow!25}0.536 & \cellcolor{green!25}0.761 & \cellcolor{yellow!25}0.682 & \cellcolor{green!25}0.911 & \cellcolor{yellow!25}0.364 \\

\textbf{$\Delta$(GPT-4o $-$ Llama)} & 
\multicolumn{2}{c|}{\cellcolor{blue!10}+0.091} & 
\multicolumn{2}{c|}{\cellcolor{blue!10}+0.164} & 
\multicolumn{2}{c|}{\cellcolor{blue!10}+0.197} & 
\multicolumn{2}{c|}{\cellcolor{blue!10}+0.079} & 
\multicolumn{2}{c}{\cellcolor{blue!10}+0.547} \\
\bottomrule
\end{tabular}}
\begin{tablenotes}
\item$^*$ A dash (--) indicates the metric is not applicable for this attribute.
\item$^*$ Bold values highlight the highest scores per metric. Colored cells distinguish GPT-4o (green) from Llama 8B (yellow).
\item$^*$ The bottom row shows the difference between GPT-4o and Llama 8B for the "Average (all)" metric.
\end{tablenotes}
\end{threeparttable}
\caption{Evaluation of the description extraction module. Expert ratings are averaged across all samples (0 = incorrect, 0.5 = partially correct, 1 = correct). According to radiologists' clinical judgment, CLEAR can accurately extract attribute-level information from free-text reports. Additionally, GPT-4o is consistently preferred over Llama-3.1-8B-Instruct, though Llama performs reasonably well, especially on \texttt{descriptive location}, and remains a low-cost, open-source option.}
\label{tab:feature}
\end{table*}

\para{\textbf{LLM-based labeler achieves substantial gains over existing labelers.}} We begin with evaluating the performance of the Label Extraction Module. As shown in \autoref{tab:classification}, our text generation-based approach (\hyperref[fig:presence-prompt]{Prompt 1}) significantly outperforms the best BERT-based labeler~\cite{smit2020combining} and the top rule-based labeler~\cite{irvin2019chexpert} across all accuracy metrics.
In identifying positive conditions, our module achieves a notable improvement in accuracy averaged over all 13 medical conditions (+15.8\%), with smaller increase on the Top 5 conditions (+12.8\%) and the full label pool (+4.1\%). This is likely because text generation models can understand the full sentence and overall report, instead of relying on token-level classification or hard-coded rules. 
Furthermore, this contextual understanding generalizes across conditions, especially for rare conditions (e.g., fracture) where BERT-based models struggle due to data imbalance, and unseen patterns (e.g., pleural other) where rule-based systems fail to capture beyond their predefined scope.
This advantage is even more evident in negative conditions, which require interpreting implicit cues (e.g., “lungs are clear”).
Our module achieves a substantial boost (+42.5\%) in average accuracy across all conditions, highlighting once again its strength in semantic understanding beyond explicit mentions.

\para{Ablation study of model scales and adaptation.} 
For identifying positive clinical findings, model scale plays a major role, with GPT-4o achieving the highest performance across all accuracy metrics.
In contrast, model adaptation strategies, including both few-shot prompting and supervised fine-tuning, have relatively limited impact compared to each base model.
This is likely because the base models already encode sufficient clinical knowledge to accurately identify positive findings, and larger model scales are more strongly related with the richness of this knowledge.
However, when it comes to negative mentions, model adaptation strategies stand out, with all metrics improving notably across scales. 
The reason is that these strategies effectively incorporate expert-derived ``side'' information, which is typically not captured by base models during pre-training, through few-shot examples or supervised training data.
Specifically, among different strategies, supervised fine-tuning consistently outperforms few-shot prompting, with average gains of 26.8\% for small models from fineuning, 7.9\% for medium models from few-shot, and 7.3\% for large models from few-shot.

\para{\textbf{LLMs, especially GPT-4o, excel at fine-grained attribute extraction.}} 
We next probe our description extraction module to assess how reliably a unified language model can handle all five fine-grained attributes (see \autoref{tab:feature}). 
Overall, GPT-4o shows strong performance across all five attributes, achieving the highest average score of 0.911 (\texttt{recommendation} average all) and a minimum of 0.733 (\texttt{severity}). 
When analyzing by task type, GPT-4o performs better on IE tasks (\texttt{location} and \texttt{recommendation}), with an average score of 0.836, particularly for attributes that involve highly formulaic language (e.g., ``follow-up imaging recommended to assess the resolution of opacity'' for \texttt{recommendation}). In contrast, it achieves a relatively lower score of 0.766 on QA tasks (\texttt{first occurrence}, \texttt{change}, and \texttt{severity}), which typically require deeper clinical contextual understanding.
In comparison, Llama-3.1-8B-Instruct (a small-scale model) shows mixed performance across attributes. In QA tasks, it captures temporal information reasonably well, scoring 0.677 for \texttt{first Occurrence} average all and 0.633 for \texttt{change}, though its interpretation of clinical findings is weaker (0.536 for \texttt{severity}). 
As for IE tasks, hallucinations significantly affect performance. But with a customized terminology list (see \autoref{tab:location-word-list}), it achieves 0.682 on \texttt{location}, the closest to GPT-4o. However, unrelated descriptive phrases (e.g., “signs of generalized fluid overload”) significantly lower \texttt{recommendation} score to 0.364.

\begin{table}[t]
\centering
\footnotesize
\begin{tabular}{lc}
\toprule
\textbf{Automated Metric} & \textbf{Corr. with Expert Scoring} \\
\midrule
\multicolumn{2}{l}{\textit{\textbf{Accuracy Metrics produced by CLEAR}}} \\
Acc. (condition averaged) & 0.894 \\
Acc. (report averaged) & 0.908 \\
Acc. (micro) & 0.915 \\
\midrule
\multicolumn{2}{l}{\textit{\textbf{Similarity Metrics produced by CLEAR}}} \\
o1-mini (micro) & 0.994 \\
ROUGE-L (micro) & 0.977 \\
BLEU-4 (micro)  & 0.811 \\
\bottomrule
\end{tabular}
\caption{Pearson correlation between CLEAR and expert scores. All of automated metrics generated by CLEAR show strong alignment with expert evaluations.}
\label{tab:cor}
\end{table}

\para{\clear aligns well with expert ratings.}
Generally, all the implementations of \clear are highly correlated with expert scoring, as shown in \autoref{tab:cor}. 
However, automated metrics are typically slightly lower than expert scores, as observed in \autoref{tab:feature}. This is because similarity metrics based on ROUGE-L and BLEU-4 prioritize exact matches against ground truth, whereas expert scoring includes a \texttt{Partially Correct} category, allowing some tolerance for clinically reasonable but not perfectly matched responses.
This distinction is further supported by the exceptionally high correlation of o1-mini scores with expert ratings, reaching 0.994. Compared to other lexical metrics, o1-mini can more effectively capture semantic and clinical alignment, making it a closer proxy to expert judgment.

\section{Related Work}
\label{sec:related_work}

\para{Lexical metrics.} Traditional word-overlap metrics such as BLEU~\cite{papineni2002bleu}, ROUGE~\cite{lin2004rouge},and METEOR~\cite{banerjee2005meteor} are commonly used in natural language generation tasks and are therefore also commonly applied to radiology report generation. 
However, these metrics fail to capture subtle semantic nuances, such as negations or synonyms, which are critical in the clinical domain.
Embedding-based metrics like BERTScore~\cite{zhang2019bertscore} improve on semantic matching but remain inadequate in capturing nuanced semantics and domain-specific medical terms, thereby missing clinically important errors.

\para{Clinical efficacy metrics.} 
To bridge the gap between surface-level fluency and clinical correctness, domain-specific metrics have been introduced. Label-based metrics such as CheXpert~\cite{irvin2019chexpert} map reports to 14 predefined clinical labels and measure classification accuracy, but their rule-based pipelines propagate annotation noise. 
CheXbert~\cite{smit2020combining} improves semantic understanding over CheXpert by fine-tuning BERT-based classifiers; however, it still lags behind recent LLMs due to the limited capacity of BERT compared to newer and more powerful language models.
More recent entity-centric methods such as RadGraph F1 \citep{jain2021radgraph}, RadGraph2~\cite{khanna2023radgraph2}, MEDCON~\cite{yim2023aci} and RaTEScore \citep{zhao-etal-2024-ratescore} capture subject–relation–object triples.
Although these approaches effectively identify and compare medical entities and their relationships, they often lack the granularity to evaluate specific attributes such as severity, temporal progression, or treatments.
To better align automatic metrics with radiologist judgments, RadCliQ~\cite{yu2023evaluating} combines BLEU, BERTScore, CheXbert similarity, and RadGraph F1 into a weighted score learned from 160 radiologist-annotated report pairs (ReXVal).
These annotations are provided at an aggregate level, quantifying the total number of clinically significant and insignificant errors without distinguishing specific clinical attributes.

\para{LLM-based metrics.} 
More recently, researchers have been using LLMs to assess radiology reports.
Several methods, including GREEN and CheXprompt, build on six categories of the clinical-error taxonomy introduced in RadCliQ.
GREEN~\cite{ostmeier2024green} tallies the number of errors and matched findings of each type and then aggregates them into a single report-level score, which limits granularity and makes it difficult to isolate specific mistakes.
CheXprompt~\cite{ZambranoChaves2025} uses GPT-4 to quantify clinically significant and insignificant errors in radiology reports, categorizing them into six predefined types.
Similarly, it focuses primarily on counting these errors without delving into the nuanced contextual attributes of each error instance. 
FineRadScore~\cite{huang2024fineradscore} takes a different route: it calculates the minimum line-by-line edits required to transform a generated report into a reference report.
While this encourages precision, it penalizes semantically equivalent but differently phrased outputs. 
RadFact~\cite{Bannur2024MAIRA2GR} decomposes each report into atomic sentences and uses LLM to determine whether each generated sentence is entailed by the reference report, which does not differentiate different types of clinical errors or severity.

\section{Conclusion}

We present CLEAR, the first clinically grounded, attribute-level evaluation framework that leverages LLMs to convert free-text radiology reports into a structured tabular format. CLEAR consists of three components: (1) a label extraction module to assess the accurate identification of medical conditions; (2) a description extraction module to evaluate the precision of condition descriptions; and (3) a scoring module to compile multi-metric evaluation results. We also introduce CLEAR-Bench, an expert-curated alignment dataset covering 6 report attributes and 13 medical conditions. Our experiments show that CLEAR can effectively identify clinical conditions, faithfully extract attribute-level information in line with clinical validation, and provide automated metrics that serve as reliable proxies for expert scoring.

\section*{Limitations}

While CLEAR provides a clinically grounded framework and demonstrates strong alignment with expert clinical assessment, it has several limitations. First, like all existing evaluation metrics, CLEAR relies solely on ground-truth reports without incorporating image information, overlooking the fact that reference reports may not fully capture all relevant findings present in the image. Future work could explore integrating image-based evaluation to better reflect clinical completeness. Second, CLEAR is built on the CheXpert label structure, which is limited in both granularity and anatomical coverage. Extending the framework to include additional specialties such as breast imaging, cardiology, and gastroenterology in the future could enhance its generalizability. Lastly, although we prioritize high-quality annotations, both the training and evaluation datasets remain relatively small due to the common tradeoff between annotation quality and dataset scale.

\section*{Acknowledgements}
We thank the anonymous reviewers for their insightful comments. We also thank Roger Engelmann from Samuel G. Armato III’s group and members of the Chicago Human+AI Lab for their valuable discussions and thoughtful feedback. This work is supported in part by NSF grants IIS-2126602 and an award from the Sloan Foundation.

\bibliography{custom}

\appendix

\section*{Appendix}

\section{Open-sourced Artifacts}
\label{sec:resource}

We release the CLEAR codebase at \url{https://github.com/ChicagoHAI/CLEAR-evaluator}
. The current version supports both open-source models via the vLLM backend and closed-source models through the Azure OpenAI API.

Our ground-truth dataset, CLEAR-Bench, along with comprehensive documentation, is publicly available on PhysioNet (\url{https://physionet.org/}
) to facilitate future research. Licensing and citation guidelines for using the dataset are detailed in the accompanying documentation.

\section{Data Annotation and Curation}
\label{sec:data}

We accessed MIMIC-CXR-JPG data by following the required steps on \url{https://physionet.org/content/mimic-cxr-jpg/2.1.0/}. We first registered and applied to be a credentialed user, and then completed the required training of CITI Data or Specimens Only Research. Data license can be found at \url{https://physionet.org/content/mimic-cxr-jpg/view-license/2.1.0/}. 

During each human annotation process, we follow a traditional paradigm: initial pilot rounds are conducted to gather user feedback, followed by formal, independent large-scale annotation, data analysis for quality control and final resolution via consensus discussion. Our annotation platform is built upon an open source data labeling tool, Label Studio~\cite{Label}.

\subsection{Label Structure Refinement}

\begin{figure}[h]
\centering
\footnotesize
\begin{tcolorbox}[colback=gray!10, colframe=gray!60, width=0.48\textwidth, boxrule=0.5pt, arc=3pt]
\textbf{MIMIC-CXR-JPG Labeling Criteria}

\vspace{4pt}
\textbf{Positive (1.0)}: The label is positively mentioned in the report and present in one or more associated images. \\
\textit{Example:} ``A large pleural effusion''

\vspace{4pt}
\textbf{Negative (0.0)}: The label is negatively mentioned in the report and should not be present in any associated image. \\
\textit{Example:} ``No pneumothorax.''

\vspace{4pt}
\textbf{Uncertain (-1.0)}: The label is either: (1) mentioned with uncertainty, so presence in the image is unclear; or (2) described ambiguously, with uncertain existence. \\
\textit{Explicit uncertainty:} ``The cardiac size cannot be evaluated.'' \\
\textit{Ambiguous language:} ``The cardiac contours are stable.''

\vspace{4pt}
\textbf{Unmentioned (Missing)}: The label is not mentioned in the report at all.
\end{tcolorbox}
\caption{4-type labeling criteria in MIMIC.}
\label{fig:mimic-label}
\end{figure}

During the interaction of pilot training, we closely work with all involved radiologists and collect a lot of valuable feedback for user experience with designed interfaces and task instruction.

After summarizing input feedback, we recognize some shared and repeatedly mentioned issues in the 4-type label structure of \href{https://physionet.org/content/mimic-cxr-jpg/2.1.0/}{MIMIC-CXR-JPG} (see \autoref{fig:mimic-label}):
(1) The ``unmentioned'' category has a high degree of overlap with other categories, particularly with ``negative'' labels. This is because radiologists often do not explicitly state negative findings in the report. However, indirect phrases such as ``Lungs are clear'' can implicitly negate a wide range of lung-related abnormalities. (2) Additionally, different radiologists have varying tendencies in labeling conditions. More conservative radiologists may lean toward assigning ``uncertain'' rather than ``positive'' labels, even when the evidence suggests a likely presence. This inconsistency introduces label noise and ambiguity, particularly when these labels are used for supervised training or evaluation purposes.

\begin{figure}[h]
\centering
\footnotesize
\begin{tcolorbox}[colback=gray!10, colframe=gray!60, width=0.48\textwidth, boxrule=0.5pt, arc=3pt]
\textbf{Our Refined Labeling Criteria}

\vspace{6pt}
\textbf{Confidently Absent}: The condition is clearly stated as not present in the report. \\
\textit{Example:} ``No pneumothorax.''

\vspace{6pt}
\textbf{Likely Absent}: The report implies the condition is likely absent, but the language is ambiguous or uncertain. \\
\textit{Example:} ``Heart size is normal though increased.''

\vspace{6pt}
\textbf{Neutral}: The report does not clearly indicate presence or absence. \\
\textit{Explicit uncertainty:} ``The cardiac size cannot be evaluated.'' \\
\textit{Ambiguous language:} ``The cardiac contours are stable.''

\vspace{6pt}
\textbf{Likely Present}: The report suggests the condition may be present, but uses uncertain or ambiguous language. \\
\textit{Example:} ``Likely reflecting compressive atelectasis.''

\vspace{6pt}
\textbf{Confidently Present}: The condition is clearly stated as present in the report. \\
\textit{Example:} ``A small right pleural effusion.''
\end{tcolorbox}
\caption{Our refined 5-type labeling criteria during expert annotation.}
\label{fig:clear-label}
\end{figure}

Therefore, we refined the original MIMIC label structure into a ``5+1'' annotation framework. The ``5'' refers to an extension of MIMIC’s original “Positive,” “Negative,” and “Uncertain” categories into five more nuanced types, as shown in \autoref{fig:clear-label}. The “+1” refers to retaining the “Unmentioned” label as a separate flag. Specifically, radiologists are asked to select one of the five labels for each condition and additionally indicate whether this label is explicitly mentioned in the report or not.

After collecting radiologist responses, we map the five types into a final three-type scheme for downstream use: “Confidently Present” and “Likely Present” are merged into “Positive,” “Confidently Absent” and “Likely Absent” into “Negative,” and “Neutral” is renamed as “Unclear.” We then proceed with inter-rater alignment checks for quality control. Notably, the “mentioned” flag is not incorporated into the final label itself but serves as a supporting indicator for data managers to differentiate between labeling disagreements due to quality issues versus differences in individual clinical interpretation. This overall process enables us to accommodate variability in radiologist judgment while maintaining high annotation quality.

\subsection{Expert-in-the-loop Dataset Curation}

We first exclude 2 cases without any ``FINDINGS'' or ``IMPRESSION'' and  30 cases labeled as “No Finding” in the radiologist annotation dataset from \href{https://physionet.org/content/mimic-cxr-jpg/2.1.0/}{MIMIC-CXR-JPG} (containing 687 studies in total). Then, we randomly select 20 cases to serve as a pilot set for initial review and refinement of the process.

\begin{figure}[t]
    \centering
    \includegraphics[width=0.5\textwidth]{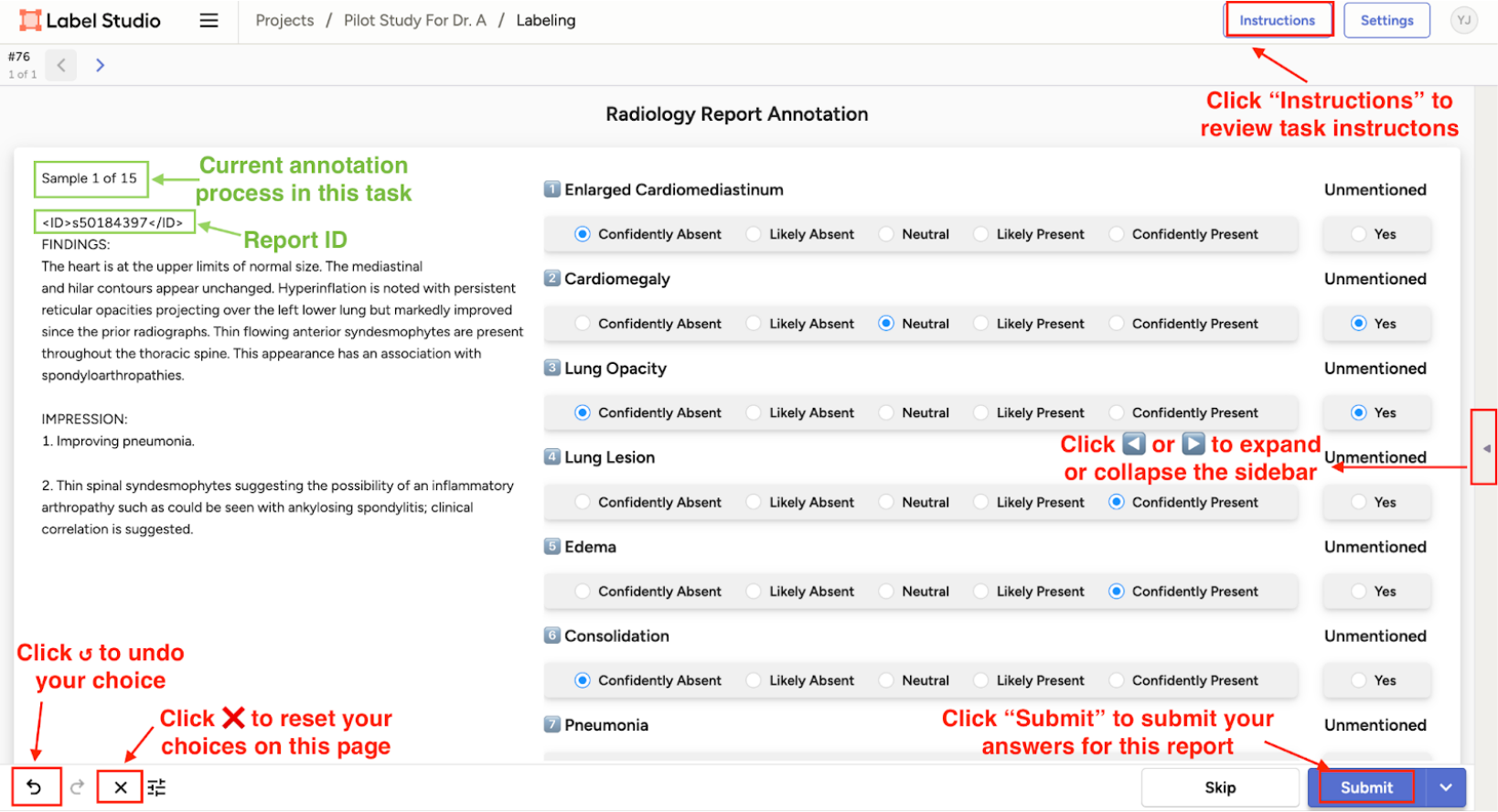}
    \caption{Interface for Label Annotation.}
    \label{fig:ui-label}
\end{figure}

We then prompt GPT-4o to generate condition labels following the same guidelines used in the original MIMIC documentation for remaining studies excluded 20 pilot cases. After identifying discrepancies between the model-generated labels and the original dataset annotations, we isolate the suspected noisy labels for further review.

For each case, we extract only the relevant report sections (FINDINGS and IMPRESSION), with no images involved, and present them to a board-certified radiologist. The radiologist independently re-annotates the report from scratch based on their clinical judgment.

During the curation, we discard 5 cases due to GPT-4o generation failures. To manage the annotation workload, we limit each review to reports with one to five mismatched conditions per case.

The full curation process took approximately one month, resulting in 550 finalized reports, each annotated with 13 condition labels.

Task instruction can be checked in \autoref{fig:ensmeble} and interface can be checked in \autoref{fig:ui-label}.

\subsection{CLEAR-Bench: Expert Ensemble}
After excluding "No Finding" cases and those already annotated in the curation stage, we selected 5 cases for pilot training and randomly sampled 100 reports from the test and validation sets of MIMIC-CXR-JPG to construct our final evaluation dataset.

Following a brief onboarding process using 5 pilot cases, we collected independent annotations from three radiologists, each labeling the 100 reports from scratch. After an initial round of majority voting, 25 reports with 32 individual condition labels in total remained unresolved. These were finalized through a single round of discussion and consensus among the experts.

The full expert ensemble workflow was completed over the course of three months, resulting in 100 fully annotated reports, each with 13 condition labels.

Task instruction can be checked in \autoref{fig:ensmeble} and interface can be checked in \autoref{fig:ui-label}.

\subsection{CLEAR-Bench: Attribute Curation}

\begin{figure}[t]
    \centering
    \includegraphics[width=0.5\textwidth]{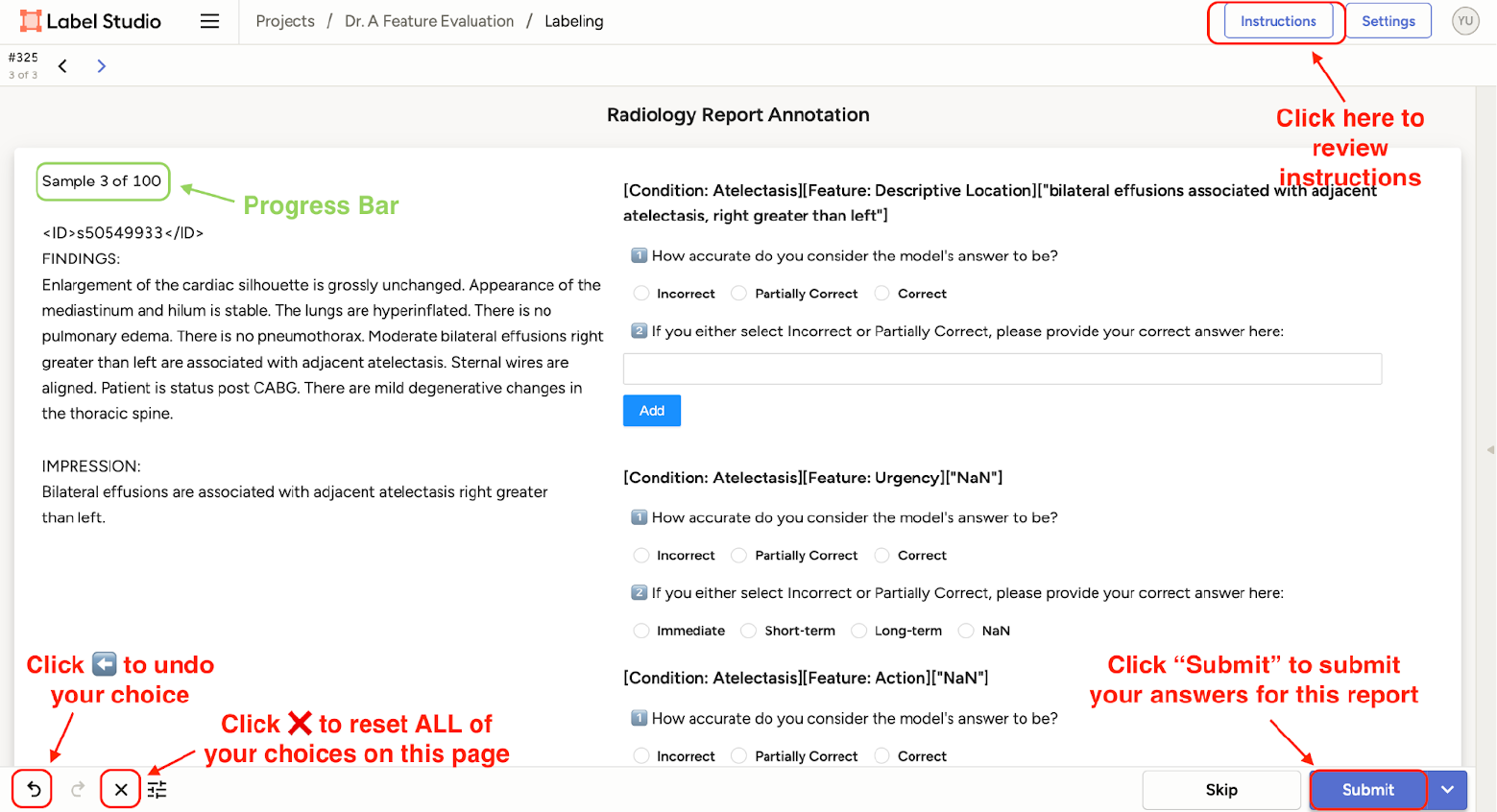}
    \caption{Interface for Attribute Curation.}
    \label{fig:ui-attribute}
\end{figure}

The blueprint for attribute design was initially inspired by the concept of an “Attribute-Value Format” proposed by Dr. Langlotz in his practical guide to writing radiology reports \citep[207]{langlotz2015radiology}. Driven by this concept, we generated a list of commonly used report attributes with the assistance of GPT-4o, and refined it through discussion with our collaborating research radiologist, who is also a co-author. Together, we determined which attributes to include, revise, or remove. During this process, we not only developed a concise yet comprehensive attribute structure but also collected useful example phrases and sentences for each attribute. These examples were later incorporated into the prompts used in the Description Extraction Module (see Appendix~\ref{sec:prompt}). The final version of the prompt set and word list was also reviewed and approved by a clinical radiologist.

We curated attributes using the same 100 studies described earlier, excluding 2 cases that lacked any positively identified conditions in expert ensemble labels. Following a round of pilot training, the formal curation process proceeded as detailed in Section~\ref{sec:clear-bench}. After collecting radiologist responses, we conducted a second round of quality control to finalize the ground-truth attributes. The full human curation process took approximately one month.

Task instructions are shown in \autoref{fig:attribute}, and the annotation interface is illustrated in \autoref{fig:ui-attribute}.

\begin{table}[t]
\centering
\small
\begin{tabular}{lc}
\toprule
\textbf{Metric} & \textbf{Correlation} \\
\midrule
BERTScore~\cite{zhang2019bertscore} & 0.27 \\
METEOR~\cite{banerjee2005meteor} & 0.49 \\
GREEN~\cite{ostmeier2024green} & 0.63 \\
GEMA-Score~\cite{zhang2025gema} & 0.70 \\
\midrule
CLEAR (GPT-4o, o1-mini) & 0.70 \\
\bottomrule
\end{tabular}
\caption{Pearson correlation with clinical judgment on ReXVal. All baseline results are reproduced from~\cite{zhang2025gema}.}
\label{tab:gema_corr}
\end{table}

\begin{table*}[t]
\small
\centering
\begin{threeparttable}
\begin{tabular}{@{}>{\arraybackslash}p{3.2cm}p{5.5cm}p{5.5cm}@{}}
\toprule
\textbf{Aspect} & \textbf{ReXVal} & \textbf{CLEAR-Bench} \\
\midrule
Report Section Used & 
Impression only & 
Full report (Findings + Impression) \\
\midrule
Annotation Scope & 
6 coarse error categories (e.g., false positive, omission, comparison) & 
13 conditions $\times$ 6 expert-curated clinical attributes \\
\midrule
Error Distribution & 
88.7\% of annotations had 0 errors; location \& change underrepresented & 
All reports selected for at least one positive finding; balanced coverage of all 6 attributes \\
\bottomrule
\end{tabular}
\caption{Comparison between ReXVal and CLEAR-Bench.}
\label{tab:rexval_clearbench}
\end{threeparttable}
\end{table*}

\section{Comparison}

\subsection{CLEAR vs. Existing LLM-based Metrics}

Notably, most existing LLM-based metrics yield a single holistic score for a (candidate, reference) pair and do not evaluate the structured, fine-grained \emph{attribute-level} information captured by our expert annotations (e.g., \emph{location} of pneumonia), which makes a like-for-like comparison between CLEAR and existing metrics impossible in Table~\ref{tab:cor}.

However, for the consideration of completeness, in this section, we constructed a naive scalarization of CLEAR by averaging its condition–attribute outputs into a single score, and evaluated it on ReXVal~\cite{yu2023rexval} following the latest GEMA protocol~\cite{zhang2025gema}. Correlation results are reported in Table~\ref{tab:gema_corr}. Under this setup, the scalarized CLEAR performs on par with GEMA-Score~\cite{zhang2025gema} and outperforms other recent semantic or clinical-efficacy metrics (e.g., GREEN~\cite{ostmeier2024green}), while still providing interpretable, condition-specific feedback that report-level metrics lack.

\begin{figure*}[t]
    \centering
    \includegraphics[width=\textwidth]{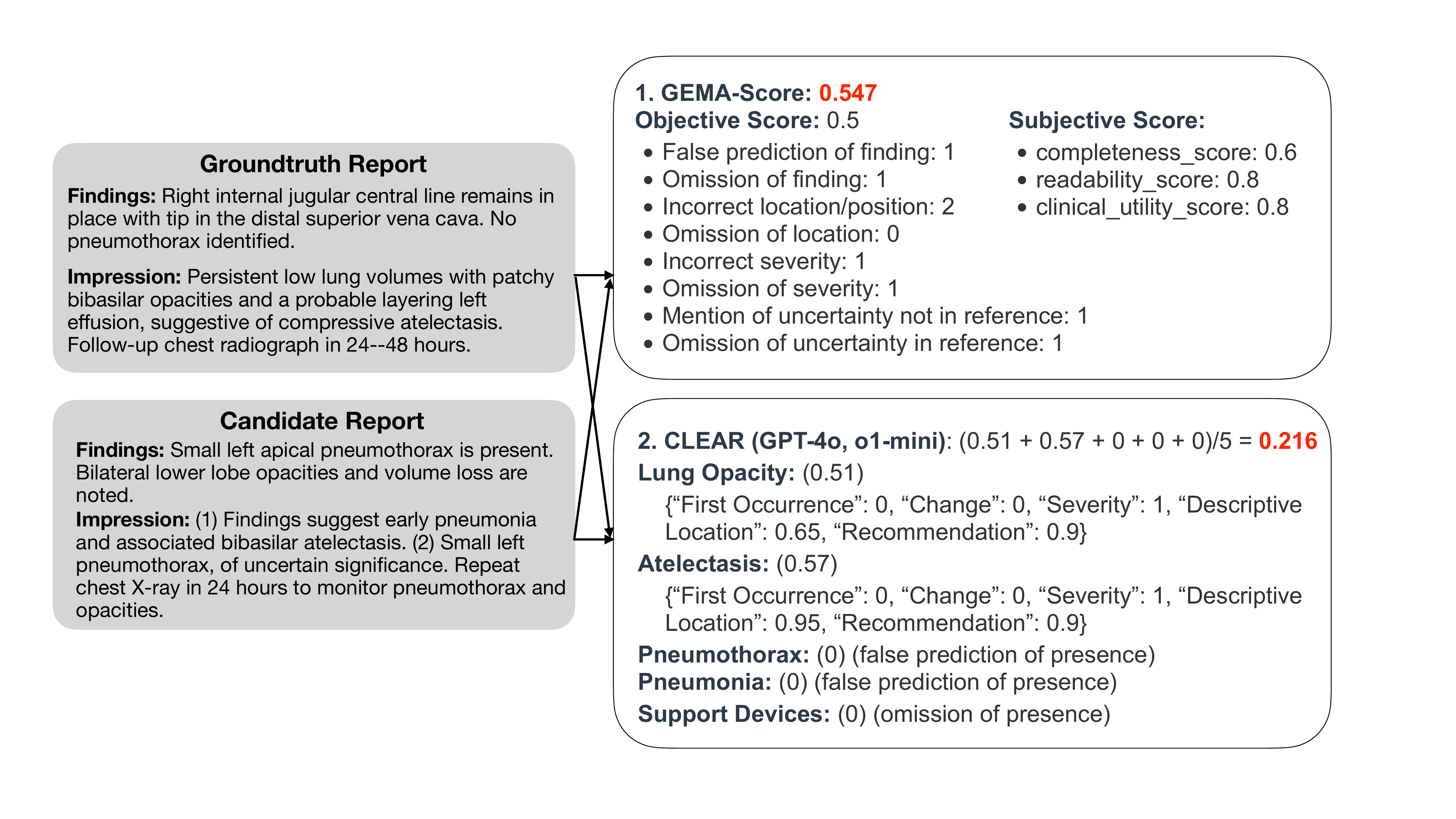}
    \caption{A case study for GEMA-Score vs.\ CLEAR (compressed version).}
    \label{fig:case-study}
\end{figure*}

We additionally present a brief case study comparing CLEAR with GEMA-Score in Figure~\ref{fig:case-study}.

Methodologically, GEMA-Score and other LLM-based radiology metrics (e.g., GREEN~\cite{ostmeier2024green}, FineRadScore~\cite{huang2024fineradscore}) follow a common paradigm: identify errors using a fixed error-category taxonomy, aggregate them into a single report-level score (e.g., F1), and assess alignment with radiologist judgments on expert-annotated sets. This paradigm is inherently constrained to single-score outputs compatible with datasets such as ReXVal~\cite{zhang2025gema}. In contrast, CLEAR is designed around a clinically interpretable, two-dimensional \emph{condition–attribute} framework. Rather than collapsing quality into one number, CLEAR produces a multi-metric “examination sheet” that explicitly indicates which conditions, attributes, and condition–attribute pairs are present or missing in generated reports.

\subsection{CLEAR-Bench vs. Existing Expert Evaluation Datasets}
Commonly used datasets such as ReXVal~\cite{yu2023rexval} are not directly compatible with CLEAR’s condition-by-attribute evaluation design. ReXVal focuses on the \emph{impression} section, lacks structured attribute annotations, and is highly sparse—over 88\% of annotator ratings indicate zero errors, with most labeled errors concentrated in a single attribute (“presence” of disease). These characteristics make it challenging to evaluate attribute-level metrics like CLEAR. 

To facilitate direct comparison where possible, we also report CLEAR’s performance on ReXVal~\cite{yu2023rexval} using the naive scalarization described above. Nevertheless, while ReXVal remains useful for coarse, report-level scoring, CLEAR-Bench is purpose-built to evaluate condition- and attribute-level fidelity and thus provides a more rigorous foundation for assessing the full capabilities of CLEAR.

Detailed comparison can be found in Table~\ref{tab:rexval_clearbench}.

\section{CLEAR: Implementation Details}
\label{sec:model}

\begin{table}[hbtp]
  \centering
  \resizebox{\linewidth}{!}{%
    \begin{tabular}{lccc}
      \toprule
      \textbf{Base Model} & \textbf{GAS} & \textbf{LR} & \textbf{Epochs} \\
      \midrule
      Llama3.1-8B-Instruct & 1 & $7.0\times10^{-6}$ & 4 \\
      Qwen2.5-7B-Instruct  & 1 & $9.0\times10^{-6}$ & 5 \\
      \bottomrule
    \end{tabular}%
  }
\caption{Hyperparameter search results. GAS denotes the number of gradient-accumulation steps, LR the learning rate, and Epochs the total training epochs.}
\label{tab:hpo_results}
\end{table}

\paragraph{Supervised finetuning details.} All fine-tuned models were obtained through supervised fine-tuning with LLaMA-Factory \cite{zheng2024llamafactory}.
To identify an optimal configuration, we developed an automated hyperparameter optimization (HPO) framework that combines five-fold cross-validation with a grid search. Learning rate, number of epoch, and gradient accumulation steps are three objects to be optimized. For learning rate, searching space is $[3.0e^{-6}, 3.0e^{-5}]$, with an interval of $2.0e^{-6}$. For epoch, searching space is $\{2, 3, 4, 5\}$. For gradient accumulation steps, searching target is $\{1,2, 4\}$.
We conduct extensive experiments to assess hyperparameters' influence.  A total of 360 models are finetuned for one base model to determine the best hyperparameter setting.
The best-performing settings, summarized in Table \ref{tab:hpo_results}, are used for all experiments reported in Table \ref{tab:classification}.
Hyperparameter optimization and model training are performed on NVIDIA A100 80G and NVIDIA H100 94G GPUs. The HPO stage takes 93 h 51 m 20 s on four A100s and 14 h 39 m 36 s on four H100s.

\paragraph{Inference details for local models.}
We serve the models locally with vLLM (0.8.5.post1) \cite{kwon2023efficient}. Inference runs with a temperature of 1e-5 and a max\_tokens of 4,096; all other sampling parameters remain at their default settings. A single NVDIA A100 80G is sufficient for inference under this setting.

\begin{table}[h]
\centering
\small
\begin{tabular}{lp{2.5cm}}
\toprule
\textbf{Model} & \textbf{Standard Pricing (per 1M Tokens)} \\
\midrule
\texttt{GPT-4o-2024-1120 (Global)} & 
Input: \$2.50 \newline
Cached: \$1.25 \newline
Output: \$10.00 \\
\texttt{o1-mini-2024-09-12 (Global)} & 
Input: \$1.10 \newline
Cached: \$0.55 \newline
Output: \$4.40 \\
\bottomrule
\end{tabular}
\caption{Standard API pricing per 1M tokens for GPT-4o and o1-mini models, based on Azure OpenAI pricing: \url{https://azure.microsoft.com/en-us/pricing/details/cognitive-services/openai-service/\#pricing}.}
\label{tab:model-pricing}
\end{table}

\paragraph{API Details}
We access OpenAI's GPT-4o (2024-11-20) and o1-mini (2024-09-12) via Microsoft’s Azure. Pricing details can be checked in \autoref{tab:model-pricing}.

\section{Template \& Terminology List}
\label{sec:prompt}

\begin{figure*}[h]
\footnotesize
\centering
\begin{tcolorbox}[colback=gray!10, colframe=gray!60, width=\textwidth, boxrule=0.5pt, arc=3pt]
Thank you very much for your support in our human annotation process! To begin with, please register at \url{https://physionet.org/content/mimic-cxr-jpg/2.1.0/} and sign the data agreement before the study. Feel free to reach us at \{\placeholder{EMAIL}\} if you encounter any issue or any questions during the process.	

\vspace{10pt}
\textbf{Overview: Task Description}

In this task, you will be extracting clinical information from \{\placeholder{NUM}\} radiology reports in total. You will not be shown the corresponding images, so you are being asked to interpret each report, as written, for the extent to which the presence of \{\placeholder{NUM}\} conditions is captured. It is important to note that some reports may have empty FINDINGS or IMPRESSION sections due to limitations in the original MIMIC-CXR-JPG database. Please follow the labeling instructions as below.

\vspace{10pt}
INSTRUCTIONS: \\
For each case, you will be presented with a single radiology report. Your objective is to choose the single most appropriate criterion among 5 options (see below) for each of the \{\placeholder{NUM}\} conditions AND note whether each condition is explicitly mentioned in the report. Please base your decisions solely on the provided report.

\vspace{10pt}
CRITERIA: \\
\{\placeholder{See \autoref{fig:clear-label}}\}

\vspace{10pt}
\textbf{Interface User Guide} \\
\{\placeholder{Account Information and Usage Tips}\}
\end{tcolorbox}
\caption{Instruction Template for Label Annotation Task}
\label{fig:ensmeble}
\end{figure*}

\begin{figure*}[h]
\footnotesize
\centering
\begin{tcolorbox}[colback=gray!10, colframe=gray!60, width=\textwidth, boxrule=0.5pt, arc=3pt]
Thank you very much for your support in our human annotation process! To begin with, please register at \url{https://physionet.org/content/mimic-cxr-jpg/2.1.0/} and sign the data agreement before the study. Feel free to reach us at \{\placeholder{EMAIL}\} if you encounter any issue or any questions during the process.	

\vspace{10pt}
\textbf{Overview: Task Description}

This curation task is to identify fine-grained features—such as location, severity, and treatment—related to specific medical conditions (e.g., edema, atelectasis, support devices) in radiology reports. You will review \{\placeholder{NUM}\} text-only reports (no X-ray images) and assess the accuracy of feature annotations generated by an AI model.  

\vspace{10pt}
Each report includes 13 predefined medical conditions, but you will only see those that were positively labeled by human annotators. As a result, the number of conditions shown per report may vary. For each positive condition, the AI extracts fine-grained details (e.g., location, severity), which you need to review. Start by marking the model’s answer as correct, partially correct, or incorrect. If it’s incorrect, enter the corrected version in the provided text box.

\vspace{10pt}
[optional] If you’d like to understand how the AI generated its responses, you can review the prompts we used at \{\placeholder{See \autoref{sec:prompt}}\}.

\vspace{10pt}
\textbf{Interface User Guide} \\
\{\placeholder{Account Information and Usage Tips}\}
\end{tcolorbox}
\caption{Instruction Template for Attribute Curation Task}
\label{fig:attribute}
\end{figure*}

\begin{figure*}[t]
\centering

\begin{promptbox}[\textbf{Prompt 1: Presence}]
\textbf{System Instruction:}

You are a radiologist reviewing a piece of radiology report to assess the presence of 13 specific medical conditions.\\

Conditions to evaluate: Cardiomegaly, Enlarged Cardiomediastinum, Atelectasis, Consolidation, Edema, Lung Lesion, Lung Opacity, Pneumonia, Pleural Effusion, Pneumothorax, Pleural Other, Fracture, Support Devices. \\

Each medical condition in the radiology report must be categorized using one of the following labels: "positive", "negative" or "unclear". The criteria for each label are:
\\
\textbullet\ \texttt{"positive"}: The condition is indicated as present in the report. \\
\textbullet\ \texttt{"negative"}: The condition is indicated as not present in the report. \\
\textbullet\ \texttt{"unclear"}: The report does not indicate a clear presence or absence of the condition. \\

The user will provide you with a piece of radiology report as input. Return your results in the following JSON format: \\
\texttt{<TASK1>\{ \\
\hspace*{1em} "Cardiomegaly": "positive"|"negative"|"unclear", \\
\hspace*{1em} "Enlarged Cardiomediastinum": "positive"|"negative"|"unclear", \\
\hspace*{1em} "Atelectasis": "positive"|"negative"|"unclear", \\
\hspace*{1em} "Consolidation": "positive"|"negative"|"unclear", \\
\hspace*{1em} "Edema": "positive"|"negative"|"unclear", \\
\hspace*{1em} "Lung Lesion": "positive"|"negative"|"unclear", \\
\hspace*{1em} "Lung Opacity": "positive"|"negative"|"unclear", \\
\hspace*{1em} "Pneumonia": "positive"|"negative"|"unclear", \\
\hspace*{1em} "Pleural Effusion": "positive"|"negative"|"unclear", \\
\hspace*{1em} "Pneumothorax": "positive"|"negative"|"unclear", \\
\hspace*{1em} "Pleural Other": "positive"|"negative"|"unclear", \\
\hspace*{1em} "Fracture": "positive"|"negative"|"unclear", \\
\hspace*{1em} "Support Devices": "positive"|"negative"|"unclear" \\
\} </TASK1>}

\vspace{1.5em}
\textbf{User Input:}

FINDINGS: \{\placeholder{findings}\} \\
IMPRESSION: \{\placeholder{impression}\}
\end{promptbox}

\captionsetup{labelformat=empty}
\caption{Prompt 1}
\label{fig:presence-prompt}

\end{figure*}

\begin{figure*}[t]
\centering

\begin{promptbox}[\textbf{Prompt 2: First Occurrence}]
\textbf{System Instruction:}

You are a radiologist reviewing a piece of radiology report to extract features for a specific condition, which was already marked as positive during the initial read of this same report.\\

Please determine from the given report (i.e., current study) whether \{\placeholder{condition}\} is being identified for the first time in current study \([\text{"current"}]\), or if the report indicates it was already present or noted in a prior study \([\text{"previous"}]\). If unmentioned, respond with \([\text{"N/A"}]\). Only choose one of the following: \([\text{"current"}]\), \([\text{"previous"}]\), or \([\text{"N/A"}]\).\\

Example answer: \([\text{"current"}]\)

\vspace{1em}
\textbf{User Input:}

FINDINGS: \{\placeholder{findings}\} \\
IMPRESSION: \{\placeholder{impression}\}
\end{promptbox}

\captionsetup{labelformat=empty}
\caption{Prompt 2}
\label{fig:occurrence-prompt}

\end{figure*}

\begin{figure*}[t]
\centering

\begin{promptbox}[\textbf{Prompt 3: Change}]
\textbf{System Instruction:}

You are a radiologist reviewing a piece of radiology report to extract features for a specific condition, which was already marked as positive during the initial read of this same report.\\

Please determine from the given report whether \{\placeholder{condition}\} is improving, stable, or worsening according to the given report. If the status is not mentioned, respond with \([\text{"N/A"}]\). If the report describes multiple statuses, respond with \([\text{"mixed"}]\). Only choose one of the following: \([\text{"improving"}]\), \([\text{"stable"}]\), \([\text{"worsening"}]\), \([\text{"mixed"}]\) or \([\text{"N/A"}]\).\\

Example answer: \([\text{"stable"}]\)

\vspace{1.5em}
\textbf{User Input:}

FINDINGS: \{\placeholder{findings}\} \\
IMPRESSION: \{\placeholder{impression}\}
\end{promptbox}

\captionsetup{labelformat=empty}
\caption{Prompt 3}
\label{fig:change-prompt}

\end{figure*}

\begin{figure*}[t]
\centering

\begin{promptbox}[\textbf{Prompt 4: Severity}]
\textbf{System Instruction:}

You are a radiologist reviewing a piece of radiology report to extract features for a specific condition, which was already marked as positive during the initial read of this same report.\\

Please determine from the given report whether \{\placeholder{condition}\} is mild, moderate, or severe according to the given report. If the status is not mentioned, respond with \([\text{"N/A"}]\). If the report describes multiple statuses, respond with \([\text{"mixed"}]\). Only choose one of the following: \([\text{"mild"}]\), \([\text{"moderate"}]\), \([\text{"severe"}]\), \([\text{"mixed"}]\) or \([\text{"N/A"}]\).\\

Example answer: \([\text{"mild"}]\)

\vspace{1.5em}
\textbf{User Input:}

FINDINGS: \{\placeholder{findings}\} \\
IMPRESSION: \{\placeholder{impression}\}
\end{promptbox}

\captionsetup{labelformat=empty}
\caption{Prompt 4}
\label{fig:severity-prompt}

\end{figure*}

\begin{figure*}[t]
\centering

\begin{promptbox}[\textbf{Prompt 5: Descriptive Location}]
\textbf{System Instruction:}

You are a radiologist reviewing a piece of radiology report to extract features for a specific condition, which was already marked as positive during the initial read of this same report.\\

Please identify the location(s) of \{\placeholder{condition}\} described in the given report. Extract and return a list of phrases that mention the anatomical location(s) \{\placeholder{location}\} specifically related to \{\placeholder{condition}\}. For each location, include any relevant descriptors {descriptor} and any associated status \{\placeholder{status}\}. \{\placeholder{note}\} If multiple phrases refer to the same location, merge them into one single entry using the most complete, informative, and non-redundant phrasing for that unique area. Format your output as one single list in the following format: \([\text{"entry-1"}, \text{"entry-2"}, \ldots, \text{"entry-n"}]\). If nothing is mentioned, return \([\text{"N/A"}]\).\\

Example answer: \\ \([\text{"left lower lobe compressive atelectasis"}, \text{"right middle lobe bibasilar atelectasis"}]\)

\vspace{1.5em}
\textbf{User Input:}

FINDINGS: \{\placeholder{findings}\} \\
IMPRESSION: \{\placeholder{impression}\}
\end{promptbox}

\captionsetup{labelformat=empty}
\caption{Prompt 5: Additional Notes: {location}/{descriptor}/{status}/{note} are a list of example key words or phrases for each condition collected from radiologists, such as (e.g., compressive, segmental, focal, terminal, peripheral, etc.).}
\label{fig:location-prompt}

\end{figure*}

\begin{table*}[t]
\centering
\renewcommand{\arraystretch}{1.2}
\resizebox{\textwidth}{!}{
\begin{tabular}{@{}p{3cm}p{4cm}p{4cm}p{4cm}p{5.5cm}@{}}
\toprule
\textbf{Condition} & \textbf{Location} & \textbf{Descriptor} & \textbf{Status} & \textbf{Note} \\
\midrule
Atelectasis & (e.g., left upper, right lower, whole lung, etc.) & (e.g., compressive, segmental, focal, terminal, peripheral, etc.) & (e.g., improving, worsening, stable, unchanged, new, etc.) & \\
\midrule
Cardiomegaly &  & (e.g., mild, moderate, severe, etc.) & (e.g., improving, worsening, stable, unchanged, new, etc.) & \\
\midrule
Consolidation & (e.g., left upper, right lower, whole lung, etc.) & (e.g., segmental, focal, terminal, etc.) & (e.g., improving, worsening, stable, unchanged, new, etc.) & \\
\midrule
Edema & (e.g., medial (near hilum), middle, lateral (peripheral), etc.) & (e.g., interstitial, alveolar, minimal, mild, moderate, severe, etc.) & (e.g., improving, worsening, stable, unchanged, new, etc.) & \\
\midrule
Enlarged Cardiomediastinum &  & (e.g., mild, moderate, severe, etc.) & (e.g., improving, worsening, stable, unchanged, new, etc.) & \\
\midrule
Fracture & (e.g., ribs, cervicothoracic vertebra, etc.) & (e.g., simple or closed, compound or open, incomplete or partial, complete, etc.) & (e.g., improving, worsening, stable, unchanged, new, etc.) & \\
\midrule
Lung Lesion & (e.g., central, peripheral, sub-pleural, entire pleural space, etc.) & (e.g., density, internal composition, shape, margin, etc.) & (e.g., improving, worsening, stable, unchanged, new, etc.) & Explicitly refer to a lung lesion (e.g., nodules, masses, infiltrates, metastases, etc.) and ignore findings unrelated to lung lesions. \\
\midrule
Lung Opacity & (e.g., left upper, right lower, perihilar, etc.) & (e.g., interstitial, alveolar, diffuse, focal, dense, ill-defined, faint, etc.) & (e.g., improving, worsening, stable, unchanged, new, etc.) & \\
\midrule
Pleural Effusion & (e.g., left, right, entire pleural space, etc.) & (e.g., subpulmonic, posterior, loculated, lobular, small, moderate, large, etc.) & (e.g., improving, worsening, stable, unchanged, new, etc.) & \\
\midrule
Pneumonia & (e.g., left upper, right lower, whole lung, etc.) & (e.g., segmental, focal, terminal, etc.) & (e.g., improving, worsening, stable, unchanged, new, etc.) & \\
\midrule
Pneumothorax & (e.g., left upper, right lower, etc.) & (e.g., simple, tension, open, etc.) & (e.g., improving, worsening, stable, unchanged, new, etc.) & \\
\midrule
Pleural Other & (e.g., left upper, right lower, entire pleural space, etc.) & (e.g., subpulmonic, posterior, loculated, lobular, diffuse, focal, etc.) & (e.g., improving, worsening, stable, unchanged, new, etc.) & Do not include findings that pertain solely to Pleural Effusion; only include findings related to other pleural abnormalities (e.g., thickening, plaques, etc.). \\
\midrule
Support Devices &  &  &  & Exclude any mention of device removal. Only include information related to existing or currently present devices. \\
\bottomrule
\end{tabular}
}
\caption{Key Words List for Location Prompt (extracted using GPT-4o, then discussed and confirmed by two radiologists)}
\label{tab:location-word-list}
\end{table*}

\begin{figure*}[t]
\centering

\begin{promptbox}[\textbf{Prompt 6: Recommendation}]
\textbf{System Instruction:}

You are a radiologist reviewing a piece of radiology report to extract features for a specific condition, which was already marked as positive during the initial read of this same report.\\

Please identify treatment(s)/follow-up(s) associated with \{\placeholder{condition}\} in the given report. Extract and return a list of phrases that only describe specific treatment(s)/follow-up(s) recommended in relation to {condition}. Do not include any phrase that merely describes the condition without any treatment/follow-up. Each treatment/follow-up should be a single entry. Format your output as a single list in the following format: \([\text{"entry-1"}, \text{"entry-2"}, \ldots, \text{"entry-n"}]\). If no action is mentioned, return \([\text{"N/A"}]\).\\

Example answer:\\
\([\text{"follow-up CT scheduled in 3 months"}, \text{"routine annual imaging advised"}]\)

\vspace{1.5em}
\textbf{User Input:}

FINDINGS: \{\placeholder{findings}\} \\
IMPRESSION: \{\placeholder{impression}\}
\end{promptbox}

\captionsetup{labelformat=empty}
\caption{Prompt 6}
\label{fig:action-prompt}

\end{figure*}

\begin{figure*}[t]
\centering

\begin{promptbox}[\textbf{o1-mini Scoring}]
\textbf{System Instruction:}

You are a radiology report comparison assistant. You will be given two lists of findings: one is the ground truth (GT), and the other is a candidate prediction (GEN). \\

Your task is to compare them and return a similarity score between 0 and 1. \\
1. A score of 1.0 means they are clinically and semantically identical. \\
2. A score of 0.0 means they are completely different or unrelated. \\
3. Partial matches should get a score in between. \\

Do not explain the score. Just output a float between 0 and 1. \\
Example answer: </SCORE>"0.8"</SCORE>

\vspace{1.5em}
\textbf{User Input:}

GT: \{\placeholder{groundtruth}\} \\
GEN: \{\placeholder{candidate}\}
\end{promptbox}

\captionsetup{labelformat=empty}
\caption{o1-mini prompt}
\label{fig:o1-mini-prompt}

\end{figure*}

\end{document}